\documentclass{article}

\usepackage[preprint]{neurips_2022}
\usepackage[utf8]{inputenc} 
\usepackage[T1]{fontenc}    
\usepackage{hyperref}       
\usepackage{url}            
\usepackage{booktabs}       
\usepackage{amsfonts}       
\usepackage{nicefrac}       
\usepackage{microtype}      
\usepackage{xcolor}         
\usepackage{algorithm}
\usepackage{algorithmic}
\usepackage{graphicx}
\usepackage{multirow}
\usepackage{amsmath}
\usepackage{amssymb}
\usepackage{mathtools}
\usepackage{amsthm}
\usepackage[capitalize,noabbrev]{cleveref}

\DeclareMathOperator*{\argmin}{arg\,min}
\usepackage{caption}
\usepackage{subcaption}
\usepackage{wrapfig}

\theoremstyle{plain}
\newtheorem{theorem}{Theorem}[section]

\newtheorem{lemma}[theorem]{Lemma}

\theoremstyle{definition}

\theoremstyle{remark}

\title{Left Heavy Tails and the Effectiveness of the Policy and Value Networks in DNN-based best-first search for Sokoban Planning}

\author{%
  Dieqiao Feng \\
  Department of Computer Science\\
  Cornell University\\
  \texttt{dqfeng@cs.cornell.edu} \\
  \And
  Carla Gomes \\
  Department of Computer Science\\
  Cornell University \\
  \texttt{gomes@cs.cornell.edu} \\
  \And
  Bart Selman \\
  Department of Computer Science\\
  Cornell University \\
  \texttt{selman@cs.cornell.edu} \\
}

\begin{document}

\maketitle

\begin{abstract}
    Despite the success of practical solvers in various NP-complete domains such as SAT and CSP as well as using deep reinforcement learning to tackle two-player games such as Go, certain classes of PSPACE-hard planning problems have remained out of reach. Even carefully designed domain-specialized solvers can fail quickly due to the exponential search space on hard instances. Recent works that combine traditional search methods, such as best-first search and Monte Carlo tree search, with Deep Neural Networks' (DNN) heuristics have shown promising progress and can solve a significant number of hard planning instances beyond specialized solvers. To better understand why these approaches work, we studied the interplay of the policy and value networks of DNN-based best-first search on Sokoban and show the surprising effectiveness of the policy network, further enhanced by the value network, as a guiding heuristic for the search. To further understand the phenomena, we studied the cost distribution of the search algorithms and found that Sokoban instances can have heavy-tailed runtime distributions, with tails both on the left and right-hand sides. In particular, for the first time, we show the existence of \textit{left heavy tails} and propose an abstract tree model that can empirically explain the appearance of these tails. The experiments show the critical role of the policy network as a powerful heuristic guiding the search, which can lead to left heavy tails with polynomial scaling by avoiding exploring exponentially sized subtrees. Our results also demonstrate the importance of random restarts, as are widely used in traditional combinatorial solvers, for DNN-based search methods to avoid left and right heavy tails.
\end{abstract}

\section{Introduction}
\label{sec:intro}

    Combinatorial search is a key domain for artificial intelligence. Unfortunately, combinatorial domains commonly have intractable theoretical complexity, such as NP-complete, PSPACE-complete, or even undecidable. In the past few decades, we have observed tremendous progress for practical problem-solving in NP-hard domains with wide applicability in, for example, circuit design \cite{hong2010qed}, hardware verification \cite{DBLP:conf/sfm/GuptaGW06}, and mathematical discovery \cite{konev2014sat}. SAT solvers based on conflict-driven clause learning can solve instances with thousands of variables and clauses in seconds, which demonstrates surprising scaling performance despite SAT being an NP-complete task \cite{silva2003grasp}.

    In contrast, practical combinatorial search in PSPACE-hard domains has remained a significant challenge. PSPACE-hard problems can be generally divided into two main categories: \emph{two-player games}, such as Go, Chess, and Amazons \cite{lichtenstein1980go, fraenkel1981computing, hearn2005amazons}, and \emph{single-agent planning problems} (a.k.a. combinatorial puzzles), such as Sokoban and problems formalized by PDDL (Planning Domain Definition Language) \cite{culberson1997sokoban, bylander1994computational}. Recent achievements in the deep learning community inspired an approach of augmenting Monte Carlo Tree Search (MCTS) with Deep Neural Networks' (DNN) guidance. AlphaGo \cite{silver2016mastering} became the first Go software to beat professional human players in 2016, and its newer and more general version AlphaZero \cite{silver2017mastering} can achieve, \emph{tabula rasa}, superhuman performance in many other challenging game domains.

    These advances naturally raise the question of whether we can build a general framework to learn different single-agent planning domains with minimum modifications. A key challenge is that hard planning instances may require intricate action sequences with hundreds of steps, and any deviation can lead to a dead-end state with no path to goal states. To address this issue, a more systematic and complete search, such as best-first search, is preferred over MCTS \cite{agostinelli2019solving, crippa2022analysis}. Traditional search methods augmented with neural network guidance have shown promising progress in planning domains \cite{shen2020learning, rivlin2020generalized, ferber2020neural, DBLP:conf/ijcai/FengGS20, feng2020novel}. These methods can solve a significant number of hard planning instances that specialized solvers cannot solve. To gain a better understanding of these approaches, we studied the interplay of the {\em policy} and {\em value networks} in DNN-based Best-First Search (DBFS) algorithms. Our experiments show the surprising effectiveness of the policy network, further enhanced by the value network, as a guiding heuristic.
    
    To further understand the phenomena, we also studied the cost distribution of DBFS. Specifically, we explored and generalized the cost distribution profiles of search methods from NP-hard domains to PSPACE-hard planning domains, and show heavy-tailed cost distributions exist ubiquitously among planning instances. For the first time, we found and characterized {\bf left heavy tails}, which are different from well-studied {\bf right heavy tails} (also abbreviated as heavy tails in the literature). Left heavy tails occur when instances become extremely hard and most runs cannot finish in a reasonable time limit. The solver needs to be ``lucky'' to occasionally hit one short run. In contrast, right heavy tails characterize mildly hard instances whose majority randomized runs have short runtime. Nevertheless, the solver can be ``unlucky'' and occasionally hit an extremely long run, which makes the expected runtime exponential. We propose an abstract search tree model that can explain the appearance of left heavy tails and provide extensive experiment data supporting our model. The experiments show the critical role of the policy network as a powerful heuristic guiding the search, which can lead to left heavy tails with polynomial scaling by avoiding exploring exponentially sized subtrees.

    Randomized combinatorial solvers use various techniques, such as randomized tie-breaking and random variable ordering, to carefully inject a \emph{controlled amount} of randomization into a deterministic search procedure \cite{crawford1994experimental, bresina1996heuristic, gomes1998boosting}. The randomization step requires careful engineering and analysis of the solver since excess randomization can hamper the effectiveness of random restarts. In our approach, we found that {\em uncertainty-aware networks} \cite{huang2017snapshot, chua2018deep, sedlmeier2019uncertainty, malinin2018predictive} can provide just the right amount of controllable randomization into a deterministic search algorithm.
    
    While we focus on Sokoban in this paper, we stress that our approach only uses the minimum domain knowledge required to describe any planning problem, namely, the state representation, the state-action transition function, and the routine deciding goal states. To make our results more general, we evaluated more than $10,000$ Sokoban instances with significant variations in the underlying combinatorial structure. 
    
    Our results also demonstrate the importance of {\em random restart strategies}, as are widely used in traditional combinatorial solvers, for Deep Reinforcement Learning (DRL) to avoid left and right heavy tails. In summary, our overall \textbf{contributions} are as follows:\\
 \textbf{(1)}  We studied the interplay of the policy and value networks in DBFS for Sokoban. Our experiments show the surprising effectiveness of the policy network, further enhanced by the value network, as a guiding heuristic for the search.\\
        \textbf{(2)}  We studied the runtime distribution on more than $10,000$ instances and propose distribution-independent statistics to quantify the heaviness of tails and effectiveness of random restarts. {\it For the first time}, we empirically studied {\bf left heavy tails} from experiment data, introduce an {\it abstract search tree model} with {\it critical nodes}, and formally show how left heavy tails can arise during the search. We explained the left heavy-tailed behavior of runtime distributions with the critical role of the policy network as a guiding heuristic. Polynomial runtime scaling can occur because the policy network helps avoid exploring exponentially sized subtrees during the search.\\
       \textbf{(3)}  We show the importance of using {\it uncertainty-aware} networks in planning and how it can add a controllable amount of randomization to a deterministic solver. We show how a {\it restart strategy} can improve DBFS's effectiveness. In particular, our experiments show for larger budgets, more frequent restarts can be more effective.

\section{Background and Related Work}
\label{sec:background}

    {\bf Sokoban as a planning domain.}\quad Sokoban is a PSPACE-complete puzzle whose goal is to push boxes into the same number of goal cells in a grid maze with walls \cite{culberson1997sokoban}. Sokoban is among the most challenging known AI planning domains. The domain remains challenging even for specialized solvers with significant human knowledge \cite{fern2011first, junghanns2001sokoban}. Due to its general search structure and hardness, we use Sokoban as our background domain throughout the paper.

    {\bf Optimal speedup of Las Vegas algorithms.}\quad Let $\mathcal{A}$ be a randomized algorithm that always outputs the correct answer when it halts but whose running time is a random variable $r^\mathcal{A}:\mathbb{Z}^+_\infty\to\mathbb{R}^+_0$. \cite{luby1993optimal} proved that when we have full knowledge about the distribution $r^\mathcal{A}$, the optimal strategy that achieves the minimum expected time required to obtain an output from $\mathcal{A}$ is to repeatedly run $\mathcal{A}$ for the same amount of time $t^\mathcal{A}$ until it halts. To calculate $t^\mathcal{A}$, let
    \[
        l(t) = \frac{1}{\sum_{x \leq t}r^\mathcal{A}(x)}(t - \sum_{x < t}\sum_{y \leq x}r^\mathcal{A}(y))
    \]
    be the expected halting time of repeatedly running $\mathcal{A}$ with time limit $t$. Define $l^\mathcal{A} = \inf_{t < \infty}l(t)$ and $l^\mathcal{A}$ is finite for any non-trivial distribution $r^\mathcal{A}$, i.e., $r^\mathcal{A}(\infty) < 1$. Let $t^\mathcal{A}$ be any finite value of $t$ such that $l(t) = l^\mathcal{A}$, if such a value exists, or
    $t^\mathcal{A}=\infty$ otherwise. \cite{luby1993optimal} also showed that when $r^\mathcal{A}$ is unknown, the universal strategy that runs $\mathcal{A}$ for time limit
    \[
        1, 1, 2, 1, 1, 2, 4, 1, 1, 2, 1, 1, 2, 4, 8, ...\footnote{https://oeis.org/A182105}
    \]
    can achieve estimated halting time $O(l^\mathcal{A}\log(l^\mathcal{A}))$ for any randomized algorithm $\mathcal{A}$. This bound is optimal among all universal strategies.

    {\bf Right heavy tails in randomized search.}\quad \cite{gomes2000heavy,gomes2005statistical} observed randomized search on SAT and CSP can exhibit right heavy tails, in particular for so-called under-constrained instances, i.e., most randomized runs on the same instance halt in a relatively short time while a non-negligible fraction of extremely long runs makes the average running time exponential. Right heavy tails have been observed in other domains such as theorem proving, planning, and scheduling \cite{meier2001heavy, cohen2018fat}. \cite{gomes2000heavy} formalized the runtime with the Pareto-L\'evy form
    \[P(X > x) \sim Cx^{-\alpha},x>0\]
    and showed random restarts can dramatically reduce the runtime variance and potentially eliminate right heavy tails. 

    {\bf Uncertainty-aware network.}\quad There is a line of research on the uncertainty of neural networks to reduce test error, provide confidence estimate, and improve model-based reinforcement learning \cite{huang2017snapshot, chua2018deep, sedlmeier2019uncertainty, malinin2018predictive}. Our method augments neural networks with Monte Carlo (MC) dropout to introduce randomization to deterministic search engines \cite{gal2016dropout}. MC dropout enables dropout layers during testing and the dropout rate can control the amount of randomization. For our experiment, the uncertainty comes from two main sources: (1) the distributional mismatch between the training and test datasets; (2) noises in the training data since the remaining distances found by specialized solvers are not optimal. See \cref{sec:formal_framework} for more details about data preparation.

\section{Formal Framework}
\label{sec:formal_framework}
    
    {\bf Policy-guided best-first search.}\quad Best-first search is an informed search algorithm, which explores a graph by expanding the most promising node chosen according to an {\bf evaluation function} $f(n)$ from the open set (search boundary nodes). $f(n)$ can use both the knowledge acquired so far while exploring the graph, denoted by $g(n)$, and a heuristic function $h(n)$, which estimates the remaining distance to the nearest goal state. Starting from the start state, best-first search gradually enlarges the current search graph by consecutively expanding a new node $n$ that minimizes the evaluation function $f(n)$ and moves $n$ to the closed set (expanded nodes). The search considers duplicate state detection and merges different nodes with the same state into a single node. Sokoban has unit cost so that $g(n)$ equals the depth of $n$. The heuristic function $h(n)$ is estimated using a value network, which is explained in detail below.

    \cite{orseau2021policy} proposed Policy-guided Heuristic Search (PHS) to further learn a policy network, which takes a state $s$ as input and outputs a vector of action probabilities $p$ with components $p(a|s)$ for each valid action $a$ of $s$. Specifically, they adapted the evaluation function to
    \[
        f(n) = \frac{g(n) + h(n)}{\pi(n)}, \pi(n) = p(s_1|s_0)\cdots p(s_m|s_{m-1}),
    \]
    where $(s_0,...,s_m)$ is the sequence of states from the root node to $n$. \cite{orseau2021policy} also proposed PHS\textsc{*}, a variant of PHS, that uses the evaluation function $f(n)=\frac{g(n) + h(n)}{\pi(n) ^ {1 + h(n) / g(n)}}$.
    
    Both PHS and PHS\textsc{*} require computing the cumulative product of probability predictions among the whole path from the root to $n$. In this paper, we use a new evaluation function that only depends on the predicted probability of the parent node of $n$:
    \[
        f(n) = \frac{g(n) + h(n)}{p(n\ |\ \textrm{parent}(n))}.
    \]
    Experiments show using this simpler evaluation function can consistently solve more instances.
    
    {\bf Data preparation.}\quad The study of the complexity and practical performance of search methods is greatly hampered by the difficulty in collecting realistic problem instance data. As an alternative, researchers heavily resort to procedurally generated instances or highly structured problem domains \cite{taylor2011procedural, guez2018investigation}. The randomly generated instances lack sufficient structure and their underlying combinatorial search space is, in some sense, too regular.

    \begin{table*}[t]
        \caption{Comparison with previous DRL works on Sokoban}
        \label{tab:data}
        \begin{center}
        \begin{small}
        \begin{sc}
            \begin{tabular}{lllll}
                \hline
                Related works & avg width & avg height & avg size & avg boxes \\
                \hline
                I2As \cite{racaniere2017imagination} & 10 & 10 & 100 & 4 \\
                PHS \cite{orseau2021policy}    & 10 & 10 & 100 & 4 \\
                \cite{DBLP:conf/ijcai/FengGS20, shoham2021solving} & 13 & 19 & 247 & 16 \\
                Our setting & 12.0 & 13.6 & 183.5 & 20.2 \\
            \end{tabular}
        \end{sc}
        \end{small}
        \end{center}
    \end{table*}
    
    To bridge this gap, we collected all the $10,871$ Sokoban instances from the Sokobano website\footnote{http://sokobano.de/en/levels.php}. All these instances were designed by different human authors in the past few decades, have a great variation in the underlying structure, serve as the benchmark for specialized solvers, and exhibit practical interest for humans to solve. The dataset is \emph{orders of magnitude larger} than the ones used in previous works on DRL and provides a great challenge for deep heuristic learning. See \cref{tab:data} for Sokoban board statistics compared with previous works. Note that the difficulty of Sokoban grows exponentially as the number of boxes increases.

    To generate supervised training data, we ran Sokolution\footnote{http://codeanalysis.fr/sokoban}, a state-of-the-art Sokoban solver, to compute ground truth plans. Sokolution can solve 8272 out of 10871 total instances given 10-minute time limit. We randomly divided the solved 8272 instances into a {\bf training set} (7435 instances) and a {\bf test set} (827 instances). For the remaining unsolved 2609 instances, we randomly sampled 200 instances and reran Sokolution with extended 2-hour time limit. 137 out of the 200 instances remained unsolved. We then collected these 137 instances as the {\bf hard set} to further study the cost distribution of instances much harder than the training instances. For each found plan $(s_0,a_1,s_1,...,a_n,s_n)$ from the start state $s_0$ to the goal state $s_n$, we generated training tuples $(s_i,l_{s_i},v_{s_i})$ with policy label $l_{s_i} = a_{i + 1}$ and remaining distance label $v_{s_i} = n - i$ as training data.

    \cite{feng2020novel} used PUSH as basic actions and achieved the state-of-the-art performance of DRL for Sokoban. In this work, we use the more elementary action MOVE. One PUSH action can be divided into two parts: 1) moving to the correct adjacent cell for pushing a box; 2) pushing a box. As a result, PUSH requires more domain knowledge of Sokoban --- the framework must compute all reachable cells from the current player position and decide which boxes are pushable. The number of valid PUSH actions can grow linearly on the number of boxes. In contrast, MOVE only consists of four actions (four directions), and way less domain knowledge is needed to decide valid moves for any state.
    Using MOVE as basic actions will generate plans that are, on average, 3-4 times longer than ones generated with PUSH actions. Most instances considered in this paper have plans containing hundreds or thousands of moves, which provides a great challenge for AI planning.

    {\bf Network architecture and training details.}\quad For each board state of height $H$ and width $W$, we created an input tensor of shape $[4\times H\times W]$ with four multi-hot feature maps encoding the player position, box positions, goal cells, and player reachable cells (ignoring all boxes), respectively. The policy head outputs a vector of length four representing the probability distribution among four moving directions. The value head outputs a single scalar representing the logarithm of the estimated remaining distance. The network consists of multiple convolutional residual blocks and each block has two extra dropout layers to introduce randomization. See
    Appendix
    for more details.
    
    The parameters $\theta$ of the deep neural network are trained with the following loss function:
    \[
        \textrm{loss} = (\log(h) - \log(v_s)) ^ 2 - \log(p_{l_s}) + c\|\theta\|^2,
    \]
    where $c$ is the weight decay parameter controlling $L_2$ weight regularization.
    
    {\bf Monte Carlo dropout.}\quad To introduce uncertainty, we used the same dropout-based architecture from \cite{gal2016dropout}. We did a grid search for the dropout ratio and found $30\%$ reached the best performance. The level of randomization is consistent and insensitive among $>100,000$ instances. We thus used dropout ratio $30\%$ as the default setting for all experiments in the following sections.

\section{The Interplay of the Policy and Value Networks}
\label{sec:interplay}
    \begin{table*}[t]
        \caption{Solver statistics of solved instances on the training and hard datasets (time limit: 10 mins).}
        \label{tab:solver}
        \begin{center}
        \begin{small}
        \begin{sc}
            \begin{tabular}{lccccccc}
                \hline
                & \multicolumn{3}{c}{Training dataset} & & \multicolumn{3}{c}{Hard dataset} \\
                \cline{2-4} \cline{6-8}
                & expanded & time & solved & nodes per sec & expanded & time & solved \\
                \hline
                Sokolution & 191436 & 310 s & 100\% & 618 & --- & --- & 0\% \\
                DBFS & 13776 & 370 s & 93\% & 37 & 18651 & 537 & {\bf 69\%}
            \end{tabular}
        \end{sc}
        \end{small}
        \end{center}
    \end{table*}
    
    \begin{table*}[t]
        \caption{Solving ratio on the test dataset with various evaluation function $f(n)$ of best-first search, depending on depth $d(n)$ (equal to $g(n)$), estimated remaining distance $h(n)$, estimated action probability $p(a|s)$, and cumulative path probability $\pi(n)$. Columns represent different search budget.}
        \label{tab:effect}
        \begin{center}
        \begin{small}
        \begin{sc}
            \scalebox{0.9}{
            \begin{tabular}{llcccccc}
                \hline
                \multirow{3}{*}{Method} & \multirow{3}{*}{$f$} & \multicolumn{6}{c}{Number of total node expansions (CPU runtime below)} \\
                \cline{3-8}
                & & $1,000$ & $2,000$ & $4,000$ & $8,000$ & $16,000$ & $32,000$ \\
                \cline{3-8}
                & & 0.5 mins & 0.9 mins & 1.8 mins & 3.8 mins & 7.4 mins & 14.5 mins \\
                \hline
                {\bf No Policy} \\
                Breadth first & $d$ & 0.32\% & 1.28\% & 1.92\% & 3.21\% & 6.73\% & 11.2\%  \\
                Greedy & $h$ & 4.17\% & 8.01\% & 12.5\% & 15.1\% & 19.2\% & 19.6\% \\
                Depth + Value & $d + h$ & 5.77\% & 10.3\% & 13.5\% & 19.9\% & 21.5\% & 25\% \\
                Depth + Value & $d + 2.0 \cdot h$ & 6.09\% & 8.97\% & 14.7\% & 17.6\% & 19.6\% & 22.8\% \\
                \hline
                {\bf With Policy} \\
                {\em Pure Policy} & $1 / p$ & {\it 28.2}\% & {\it 32.1}\% & {\it 36.9}\% & {\it 40.4}\% & {\it 42.9}\% & {\it 44.6}\% \\
                PHS & $(d + h)/\pi$ & $15.7\%$ & 19.9\% & 23.7\% & 26.3\% & 29.2\% & 31.4\% \\
                PHS\textsc{*} & $(d + h)/\pi^{1 + h/d}$ & 28.5\% & 31.4\% & 38.5\% & 40.1\% & 44.9\% & 46.2\% \\
                Policy + Greedy & $h / p$ & 31.7\% & 32.4\% & 37.5\% & 38.8\% & 41.0\% & 41.3\% \\
                Ours & $(d + h) / p$ & {\bf 32.4\%} & {\bf 34.3\%} & 38.8\% & {\bf 43.3\%} & {\bf 46.2\%} & {\bf 50.0\%} \\
                Ours & $(d + 2.0 \cdot h) / p$ & 31.7\% & 34.0\% & {\bf 39.4\%} & 42.0\% & 45.8\% & 48.1\% \\
            \end{tabular}
            }
        \end{sc}
        \end{small}
        \end{center}
    \end{table*}

    In this section, we first show the performance comparison between DNN-guided search and specialized solvers, and further show how different evaluation functions affect the overall performance.
    
    {\bf Solver statistics.}\quad \cref{tab:solver} shows solver statistics of Sokolution, a state-of-the-art solver of Sokoban, and DBFS with the evaluation function $f(n) = \frac{g(n)+h(n)}{p(n\ |\ \textrm{parent}(n))}$. We used 8 cores of Xeon 6154 CPUs for profiling both solvers and ran networks on the CPU mode for a fair comparison. DBFS expands significantly fewer nodes per second than Sokolution (about a factor of 17) due to the heavy cost of heuristic evaluation. Nevertheless, {\em given 10 minutes, DBFS can solve 69\% hard instances that Sokolution cannot solve even given a 2-hour time limit}. So, the trained deep net provides much superior search guidance than the hand-crafted guidance in Sokolution.
    
    {\bf Effectiveness of policy and value networks.}\quad \cref{tab:effect} shows experiment results for different choices of the evaluation function. Term $p$, $h$, and $d$ represent policy, value, and depth information, respectively. As shown in the table, {\em the policy heuristic has a significantly larger impact than the value heuristic.} Specifically, the table shows that even the Pure Policy (using only the $1/p$ term, i.e., inversely proportional to the policy prediction) significantly boosts performance compared to all value (the $h$ term) heuristics-based search strategies without the policy guidance. (See the rows above ``Pure Policy'' in \cref{tab:effect}.)
    With extra properly added depth and value terms, the performance of Pure Policy can further increase to obtain our best strategies.
    
    \begin{wraptable}{r}{0.4\textwidth}
        \caption{Dead-end detection accuracy}
        \label{tab:end}
        \begin{center}
        \begin{tabular}{lcccccc}
            \hline
            & Train & Test & Hard \\
            \hline
            Policy & {\bf 93\%} & {\bf 81\%} & {\bf 68\%}\\
            Value & 41\% & 38\% & 37\%
        \end{tabular}
        \end{center}
    \end{wraptable}
    To further study why the policy network is more effective, we studied the performance of both networks in detecting dead-end states. Detecting dead-end states is crucial to avoid exponential run times because the search has to recover from exploring those states. In particular, we randomly sampled 2000 board states from each dataset. For each state, we used Sokolution to detect dead-end successor states. The policy/value network is considered to successfully detect a state if it predicts a higher/lower policy/value to the child on the ground truth plan than any other dead-end child. \cref{tab:end} shows that the policy network significantly outperforms the value network at detecting dead-ends, providing better search guidance.

\section{Analysis of Left Heavy Tails}
\label{sec:analysis}

    \begin{figure*}[t]
        \begin{center}
            \includegraphics[width=\textwidth]{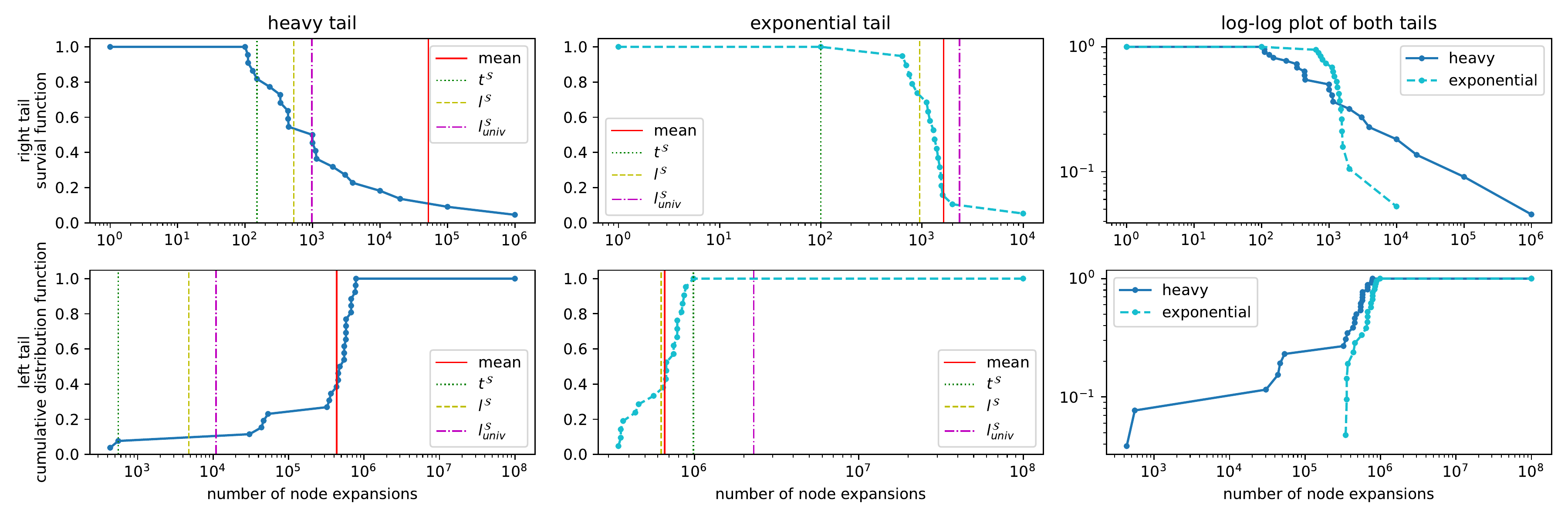}
        \end{center}
        \caption{Each subplot shows runtime statistics of DBFS with MC dropout augmented on Sokoban instances. Each curve represents multiple runs on the same instance (instances differ for different curves). We compare the runtime sample mean, optimal sample restart time $t^\mathcal{S}$, expected sample total runtime with restart $l^\mathcal{S}$, and expected total runtime of the universal strategy $l_{univ}^\mathcal{S}$ as defined in \cref{sec:analysis}.}
        \label{fig:tails}
    \end{figure*}
    
    \begin{figure*}[t]
        \begin{center}
            \includegraphics[width=\textwidth]{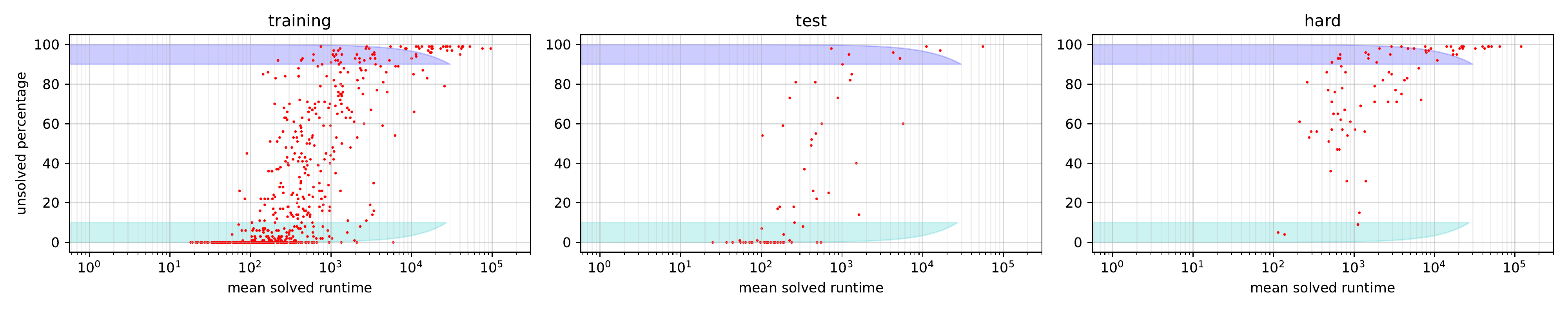}
        \end{center}
        \caption{Subplots show results of randomized DBFS for the training, test, and hard datasets. Each red dot represents a single instance. We perform 200 randomized searches on each instance with a maximum of $300,000$ node expansions. The X-axis is the average runtime of all solved runs and the Y-axis is the unsolved percentage. Purple areas represent instances with left heavy tails while cyan areas represent right heavy tails.}
        \label{fig:both_tails}
    \end{figure*}
    
    In \cref{sec:interplay}, $31\%$ instances of the hard dataset remained unsolved. We increased the search budget and studied the runtime distributions of these challenging instances and found they possess heavy tails on the left-hand side. In this section, we study this behavior in further detail. 
    
    {\bf Sample statistics about the heaviness of tails.}\quad Luby's optimal restart strategy requires full knowledge about the runtime distribution $r^\mathcal{A}$ of a randomized algorithm $\mathcal{A}$. Here we introduce sample statistics that can approximate the theoretical optimal values. Specifically, for each planning instance, we performed multiple randomized searches on it with maximum search budget $\mathcal{M}$ and collected a runtime sample $\mathcal{S}$. The runtime of failed searches was capped as $\mathcal{M}$. We approximated the optimal sample restart time $t^\mathcal{S}$ and expected sample total runtime with restart $l^\mathcal{S}$ as
    \begin{align*}
        t^\mathcal{S} &= \argmin_{u\in\mathcal{S}}\frac{u\cdot|\mathcal{S}|}{|\{v|v\in\mathcal{S}\textrm{ and }v\leq u\}|}, \\
        l^\mathcal{S} &= \frac{t^\mathcal{S}\cdot |\mathcal{S}|}{|\{v|v\in \mathcal{S} \textrm{ and } v\leq t^\mathcal{S}\}|}.
    \end{align*}
    Let $T=(1,1,2,1,1,2,4...)$ be the time limit sequence of the universal strategy. To approximate the expected total runtime $l_{univ}^\mathcal{S}$ of the universal strategy on $\mathcal{S}$, let $a_i$ be the expected total runtime of applying $(T_i,T_{i+1},T_{i+2},...)$ on $\mathcal{S}$ and we have the constraint
    \[
        a_i = T_i + \frac{a_{i+1}\cdot|\mathcal{S}|}{|\{v|v\in\mathcal{S}\textrm{ and }v >  T_i\}|},\quad l_{univ}^\mathcal{S}=a_1.
    \]
    We calculated $a_i$ until the first $i$ such that $T_i \geq \mathcal{M}$ and set the remaining $a_i$ to zeros.

    \cref{fig:tails} shows runtime statistics for different types of tails. Both left and right heavy tails exhibit orders of magnitude reduction of $l^\mathcal{S}$ and $l^\mathcal{S}_{univ}$ over the runtime sample mean, which demonstrates the benefit of using random restarts. For exponential tails, the expected sample total runtime with restart is very close to the sample mean, and the universal strategy even has a negative effect.
    
    To separate the two types of heavy tails, for each instance, we compared the average runtime of solved randomized runs v.s. the unsolved ratio, as shown in \cref{fig:both_tails}. We added further restrictions that a right/left heavy tail requires the unsolved ratio to be less/greater than 10\%/90\%. For experiment budget concerns, we only plotted 10\% randomly sampled instances from the training and test datasets. \cref{fig:both_tails} shows the training dataset has a similar number of left and right heavy-tailed instances, with most instances not showing the heavy-tailed behavior. For the hard dataset, the percentage of right heavy tails decreases significantly, with more instances shifting to the top side of the figure and entering the left heavy-tailed area. We hypothesize that left heavy tails occur more frequently when the underlying combinatorial structure becomes harder. Though random restarts can eliminate heavy tails on both sides, left heavy tails provide further intuitions and implications for curriculum learning for DRL. In particular, they can benefit from a distributed solving procedure in which any solution found with one process can be shared and learned by the curriculum framework \cite{weng2020curriculum, narvekar2018learning, narvekar2020curriculum}.
    
    \begin{figure*}[t]
        \begin{center}
            \begin{minipage}{0.3\textwidth}
                \centering
                \includegraphics[width=\textwidth]{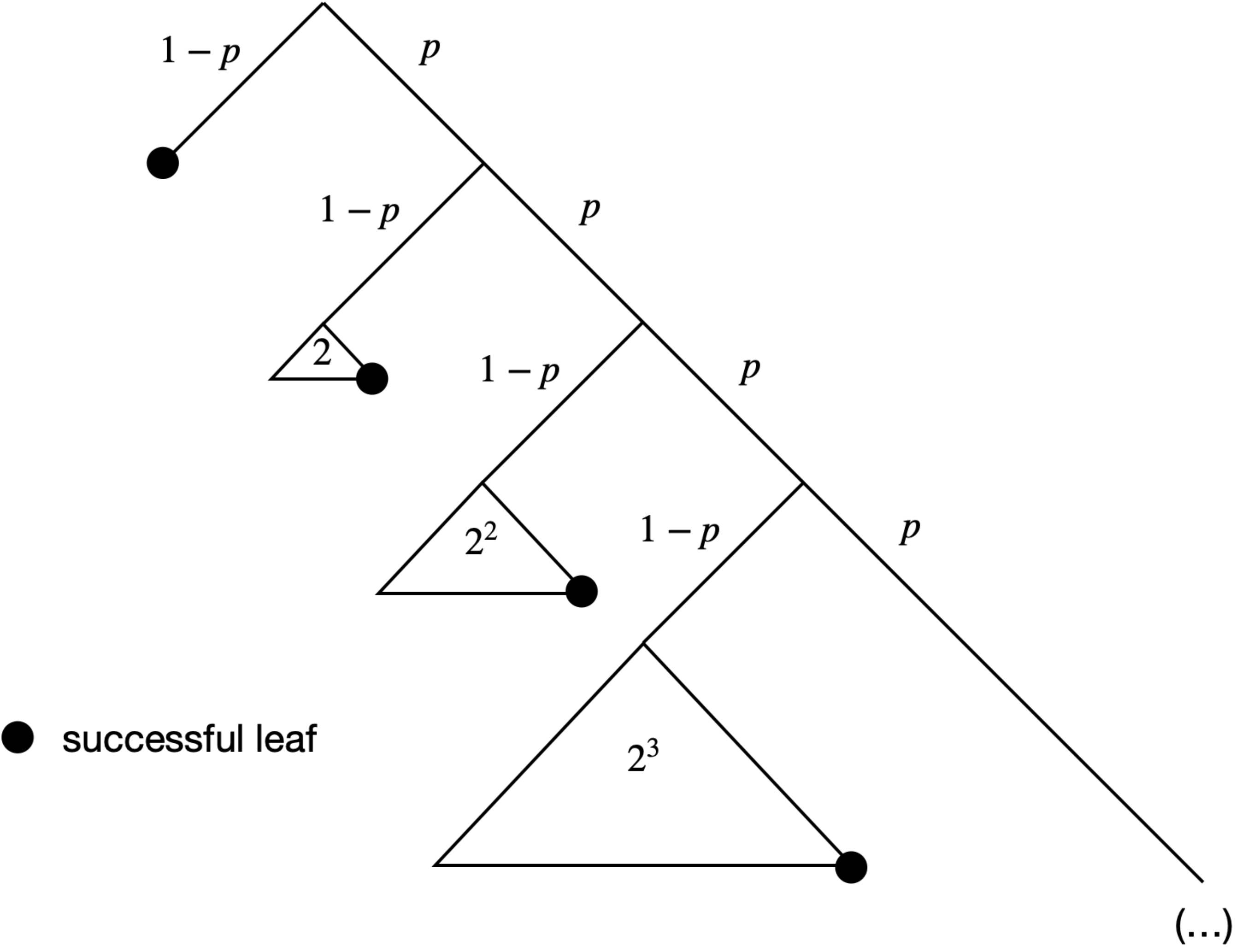}
            \end{minipage}\hspace{0.2\textwidth}%
            \begin{minipage}{0.3\textwidth}
                \centering
                \includegraphics[width=\textwidth]{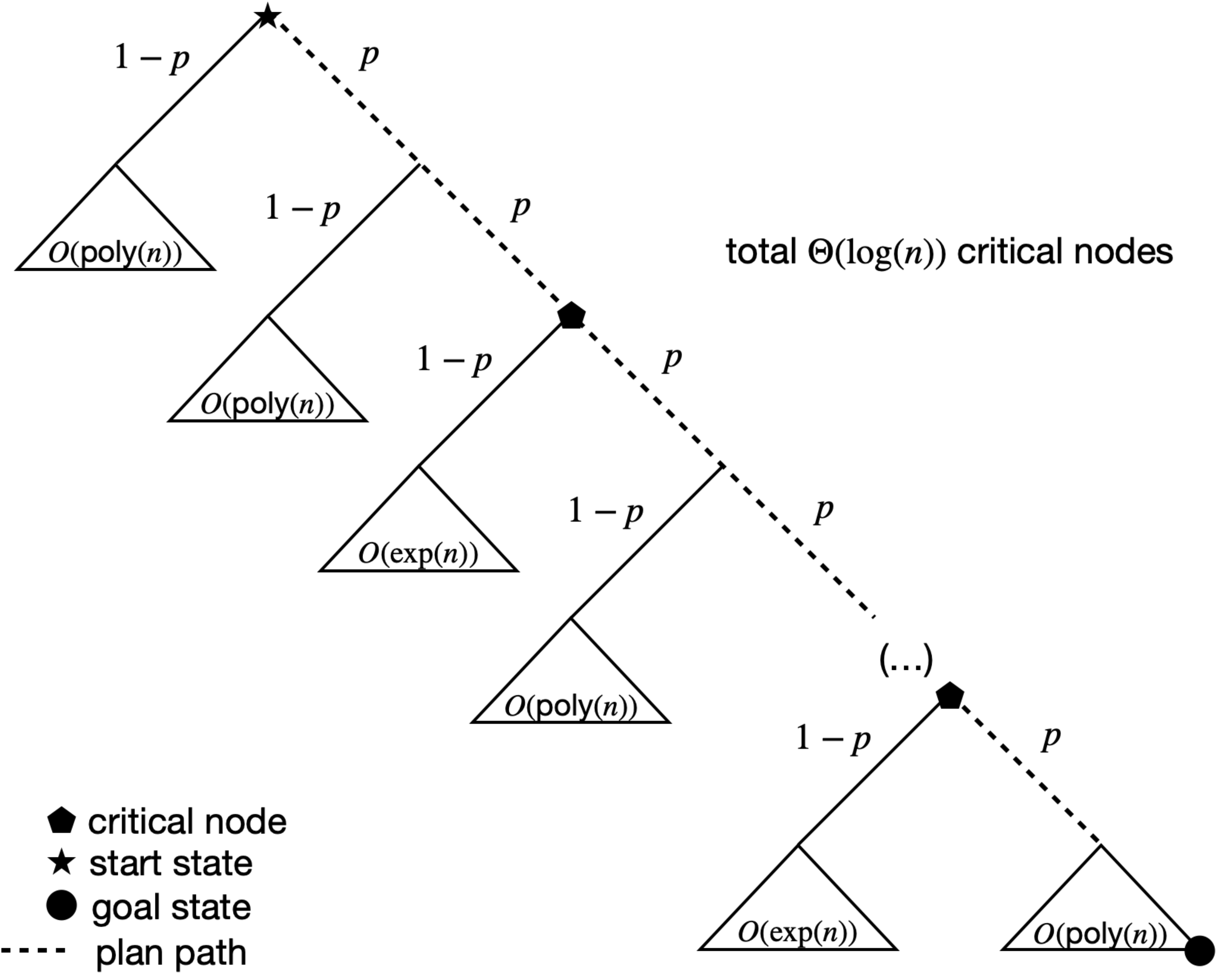}
            \end{minipage}
        \end{center}
        \caption{Left panel: imbalanced tree model for right heavy tails. Right panel: our proposed model for left heavy tails. In the left model, $p$ is the constant probability of missing a backdoor. In the right model, $p$ is the constant probability of picking the right action to the goal.}
        \label{fig:model}
    \end{figure*}
    
    {\bf Abstract search tree model.}\quad \cite{chen2001formal} proposed an imbalanced tree model to empirically explain right heavy tails. Here we propose an abstract tree model to empirically characterize left heavy tails. See \cref{fig:model} for the description for both models.

    Search models for planning problems differ from ones for SAT and CSP. For NP domains, the number of unassigned variables is fixed with $O(n)$ where $n$ is the problem size, and the search can assign these variables in any order. For planning domains, the search must assign actions in order from the start state to a goal state with potentially maximum exponential length. Our proposed model hypothesizes the existence of $O(log(n))$ {\bf critical nodes} from which a wrong child node expanded by the search will result in extra exponential search space. As shown in follows, our model does not depend on the actual choice of $p$ as long as $p$ is a constant value in the range $(0, 1)$.

    \begin{theorem}
        \label{thm:one}
        {\bf (The abstract tree model has exponential runtime almost surely).} When restricting plans to polynomial length on the input size $n$, the probability that the abstract tree model has exponential runtime converges to 1 as $n$ goes infinite. \textup{The proof is deferred to
        Appendix
        }
    \end{theorem}
    \begin{theorem}
        \label{thm:two}
        {\bf (Restart achieves polynomial expected halting time).} The optimal expected halting time with restart $l^\mathcal{A}$ and expected halting time using the universal strategy $l^\mathcal{A}_{univ}$ are both $O(\textrm{poly}(n))$. \textup{The proof is deferred to
        Appendix
        }
    \end{theorem}
    
    \cref{thm:one} states the runtime of the model is exponential on the input size $n$ almost surely, and \cref{thm:two} shows that the cost distribution has polynomial estimated runtime with restarts. These two theorems combined show the occurrence of left heavy tails. For most nodes on the plan, the deep neural network either provides accuracy heuristics to expand the correct child node or makes a small error of preferring one wrong child node. As long as the error is small, the search can recover from it with extra polynomial steps since the evaluation function penalizes deeper nodes. For the $O(\log(n))$ critical nodes, the error of predicted heuristics is so large that exponential search is needed to jump out from the local search space.
    
    \begin{figure*}[t]
        \begin{center}
            \includegraphics[width=0.9\textwidth]{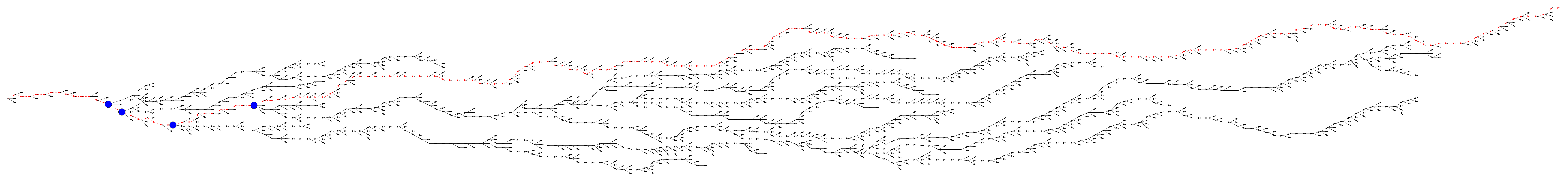}
            \includegraphics[width=0.9\textwidth]{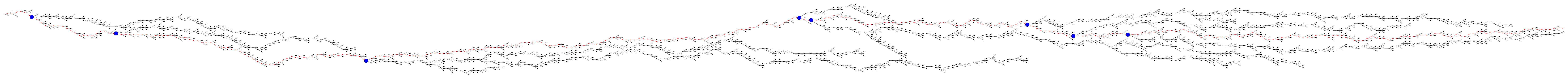}
        \end{center}
        \caption{Two typical search graphs of DBFS. We removed graph edges to form a spanning tree for a clearer illustration. The search graph was built from the left to the right. The red dashed path is the plan leading from the start state to one goal state, while blue circles are \emph{critical nodes}, which have exponential sub-search tree underneath without a near goal state.}
       \label{fig:graph}
    \end{figure*}
    
    {\bf Structure of real search graphs.}\quad To empirically examine our abstract model, we plotted two typical search graphs generated by DBFS as shown in \cref{fig:graph}. We found that most nodes actually {\bf do not} incur search. Networks' heuristics either directly lead the search to the correct child node, or only a small wrong subtree (less than 5 nodes) is expanded. \emph{Critical nodes} (labeled as blue circles) are extremely rare in the search graph. However, when encountering such critical nodes, the search expands a large subtree with no near goal.

    Whether AI planning systems can discover {\bf macro action routines}, i.e., a sequence of algorithmic actions to perform a sub-goal, has a great interest for researchers. To make a long plan, e.g., prove a hard mathematical theorem, humans usually only make some \emph{critical choices} of lemmas and schemes, and fill the remaining parts of the proof with little reasoning. Indeed, the number of required crucial intermediate lemmas to prove a mathematical theorem is quite small, even for challenging open problems. However, extensive and profound reasoning, search, and enumeration are needed to find such lemmas. The small number of critical lemmas compared with the long proof length reflects the prototype of such search graphs. In our experiment setting, we use MOVE as basic actions. To perform a real PUSH, the search algorithm must compute all reachable cells from the player's position and calculate the shortest path to the cell adjacent to the box to push. Such a long sequence of moves before performing an actual push can be viewed as a {\bf routine}. As shown in \cref{fig:graph}, the algorithm can perform a long chain of moves with little local search, which suggests that macro action routines might be {\bf implicitly} learned as a part of neural networks' heuristics.

    {\bf Relation to backdoors.}\quad The proposed model has a close relation to backdoors of typical case complexity. To explain why solvers scale so well in areas such as planning and finite model-checking, \cite{williams2003backdoors} examined various benchmarks and identified that for most practically solvable problem instances, after assigning values to \emph{logarithmic} variables, the remaining problem instance quickly becomes polynomially solvable by propagating constraints. This result illuminates the prototypical patterns of the structure causing the empirical behavior observed in the International Planning Competitions benchmarks \cite{vallati20152014, cohen2018fat, meier2014randomization}.

    \begin{figure}
        \centering
        \begin{subfigure}{.45\textwidth}
            \centering
            \includegraphics[width=\linewidth]{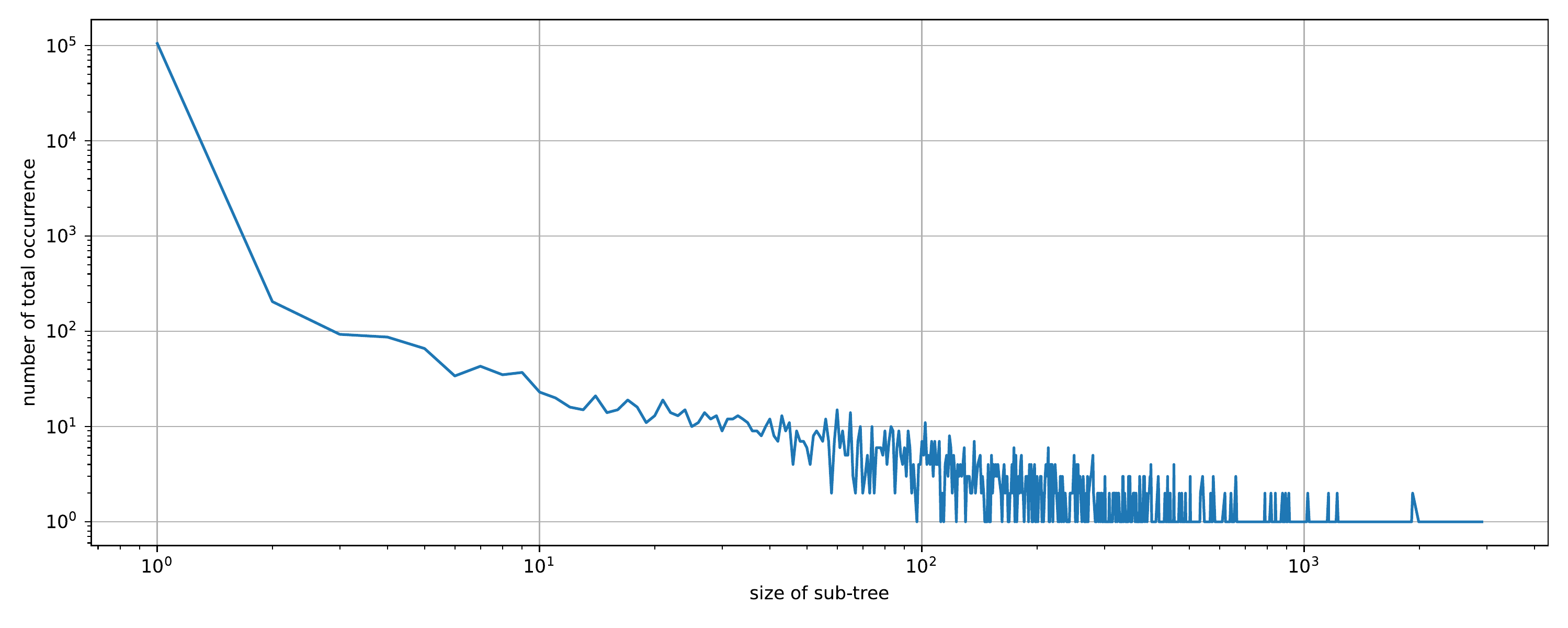}
            \caption{Subtree size v.s. total number of occurrence for 300 randomly sampled instances (in log-log scale).}
            \label{fig:stats}
        \end{subfigure}%
        \hspace*{\fill}
        \begin{subfigure}{.45\textwidth}
            \centering
            \includegraphics[width=\linewidth]{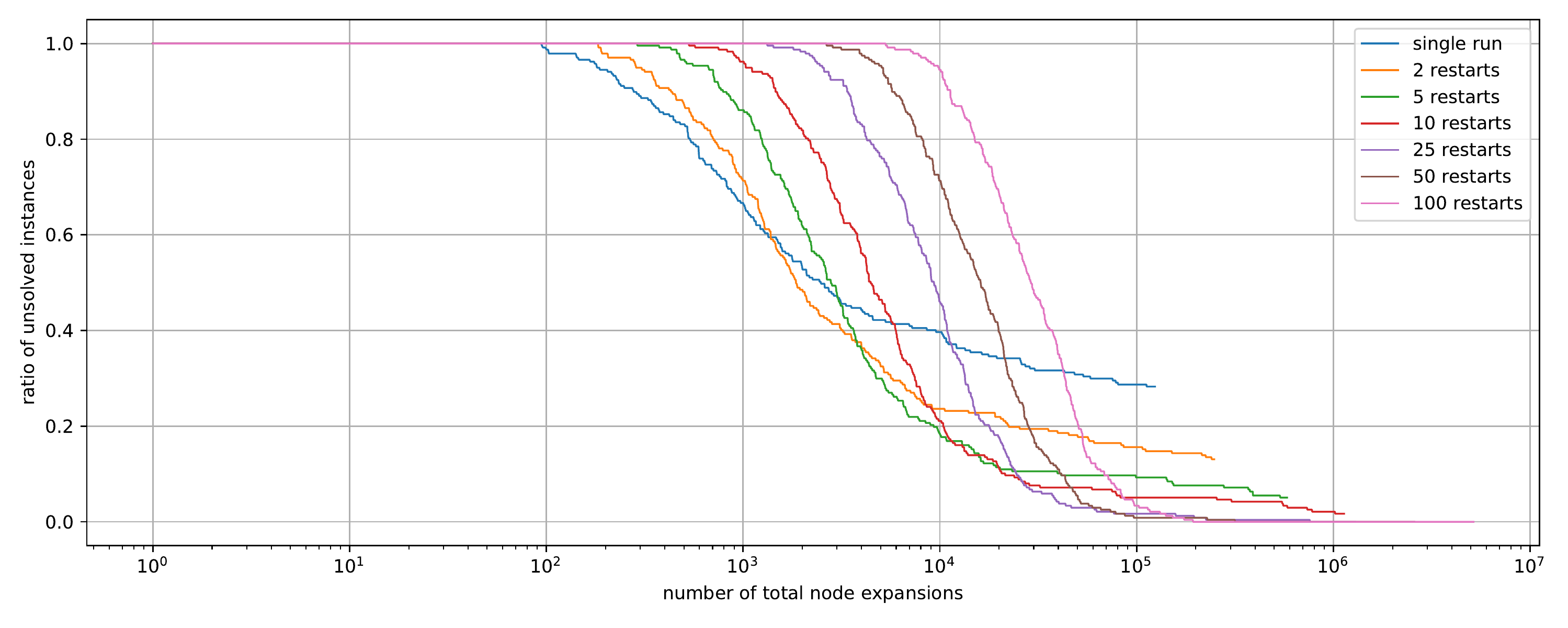}
            \caption{Unsolved fraction of instances with or without restarts. Given a fixed amount of search budget (number of nodes allowed to expand), $n$-restart means to evenly divide the search budget into $n$ individual runs and union their solved instance sets as the final result.}
            \label{fig:restart}
        \end{subfigure}
    \end{figure}
    
    {\bf More experiment data about the abstract tree model.}\quad To better illustrate the occurrence probability of different sizes of sub-search space, we randomly sampled 300 solved instances from the test dataset and counted the number of dead subtrees (no near goal exists) for each size. As shown in \cref{fig:stats}, in log-log scale, the size of subtrees and the occurrence have a near-linear correlation. The figure empirically confirms our abstract tree model that critical nodes that incur extra exponential search space have logarithmic occurrence. And it shows how left heavy tails occur in practical planning solving.

    \cite{hoffmann2006structure} explained SATPLAN, classical planning as SAT, with backdoor models. In particular, after finding and assigning logarithmic variables, the remaining problem solving becomes polynomial. In contrast, identifying these backdoors is not required for expand-style search algorithms in planning. Left heavy tails are not only caused by the underlying structure of practical instances but also affected by the mysterious generalization power of neural networks. It is an interesting future research direction to understand the surprising scaling performance of various heuristics, from conflict-driven clause learning for SAT solving to DNN-based search methods.

    {\bf Solving more instances with random restarts.}\quad The theory of heavy-tailed cost distribution suggests that a sequence of short runs instead of a single long run make better use of a fixed amount of computational budget. We explored this idea by considering a fixed number of total expanded nodes allowed for the search. \cref{fig:restart} shows the probability of \emph{not} solving the instances for the test dataset. The failure rate of a single run drops the fastest for a small amount of computational budget. With more total expanded nodes, random restarts gradually achieve better performance. Specifically, to solve more instances, the solver needs to increase the total number of compute cycles. When doing so, the figure shows there comes a point where more frequent restarts are more effective. For example, with a budget of around 2,000 nodes, the strategy with 2 restarts becomes more effective than no restarts. At around 5,000 nodes, 5-restart becomes more effective than 2-restart. So, to solve a larger fraction of hard instances, more frequent restarts become more effective.

\section{Conclusions}
    We studied the use of policy (action selection) and value (remaining distance estimate) functions as well as randomization methods for solving hard planning instances using best-first search. Our experiments show the remarkable effectiveness of the policy network and random restarts for the search. The value network provides additional global search guidance.
    
    We show that {\it uncertainty-aware} networks provide an effective way to introduce randomization into the search process leading to increased efficiency. Our runtime distribution results show {\it heavy-tailed distributions} with tails on both the left and right-hand sides. Left heavy tails have not been observed in combinatorial search before. We also introduce an abstract computational model that explains left heavy tails. Finally, we show how {\it random restarts} can improve the overall search effectiveness. With larger search budgets, restarts can be increasingly effective.

\clearpage
\bibliography{neurips_2022}


\clearpage

\appendix
\onecolumn
\section{The Imbalanced Tree Model Proofs}
\subsection{Proof of \cref{thm:one}}
\label{proof:one}

\begin{lemma}
    \label{lemma_one}
    The abstract tree mode has polynomial runtime if and only if all critical nodes pick the right child.
\end{lemma}

Proof of \cref{lemma_one}.\quad Necessity comes from the structure of the model and any picking of left child will immediately result in exponential search space. For sufficiency, by the assumption, the depth of the tree model is bounded by $O(\text{poly}(n))$, and the final sub-search space with all right child picking is also $O(\text{poly}(n))$. Then we have the total runtime is $O(\text{poly}(n)) + O(\text{poly}(n)) = O(\text{poly}(n))$. \hfill Q.E.D.

Proof of \cref{thm:one}. \quad By \cref{lemma_one}, we have
\begin{align*}
\lim_{n\to\infty}\Pr(\text{model has poly runtime}) &= \lim_{n\to\infty}p^{\Theta(\log(\text{poly}(n)))} \\
&\leq \lim_{n\to\infty}p^{C\cdot\log(\text{poly}(n))} && (\textrm{for some constant $C$}) \\
&= 0 && (\textrm{$0 < p < 1$ as assumption}).
\end{align*}
As a conclusion, $\lim_{n\to\infty}\Pr(\textrm{tree mode has exponential runtime}) = 1$.\hfill Q.E.D.

\subsection{Proof of \cref{thm:two}}
\label{proof:two}
Proof.\quad We can assume there exists at least one non-critical node on the plan (if all tree nodes are critical, we can augment an extra non-critical node to the right child of the deepest critical node without affect other properties of the tree model). Let $u$ be the shallowest non-critical node and $d$ be the depth of node $u$. We set the restart time $t^\mathcal{A}$ to be the size of the left subtree of $d$. $t^\mathcal{A}$ is $\textrm{poly}(n)$ by the definition of the model. Let $q=p^d(1-p)$. So the expected runtime
\begin{align*}
    \lim_{n\to\infty}l^\mathcal{A} &= \lim_{n\to\infty}(d + O(\textrm{poly}(n)))\sum_{k=0}^{\infty}q\cdot(1-q)^k\cdot (k+1) \\
    &= \lim_{n\to\infty}(d + O(\textrm{poly}(n))) \cdot \frac{1}{q} \\
    &= O(\textrm{poly}(n)).
\end{align*}

By the result of \citet{luby1993optimal}, $l^\mathcal{A}_{univ}=O(l^\mathcal{A}\log(l^\mathcal{A})) = O(\textrm{poly}(n))$.\hfill Q.E.D.

\section{Network Architecture and Training details}
\label{sec:network}
\subsection{Network Architecture}
A single input tensor of board states has shape $[4\times H\times W]$ and a batch of board states can have different heights and widths. For each batch, we take the maximum $H$ and $W$ of all state tensors as the batch height and width, and zero-pad the empty cells.

The network consists of a single convolution block followed by 16 residual blocks.

The convolutional block applies the following modules:
\begin{enumerate}
    \item A convolution of 128 filters of kernel size $3\times 3$ with stride $1$
    \item 2D batch normalization \citep{ioffe2015batch}
    \item A ReLU nonlinearity
\end{enumerate}
Each residual block applies the following modules sequentially to its input:
\begin{enumerate}
    \item A channel-wise dropout layer with probability 30\% of a channel to be zeroed.
    \item A convolution of 128 filters of kernel size $3\times 3$ with stride $1$
    \item 2D batch normalization
    \item A Relu nonlinearity
    \item A channel-wise dropout layer with probability 30\% of a channel to be zeroed.
    \item A convolution of 128 filters of kernel size $3\times 3$ with stride $1$
    \item 2D batch normalization
    \item A skip connection that adds the input to the block
    \item A Relu nonlinearity
\end{enumerate}
The output of the residual tower is then fed into two independent heads for computing the policy and value. Both heads contain an extra residual block followed by a fully connected layer. The policy head outputs a vector of size 4 and the value head outputs a single scalar.
\label{sec:arch}
\subsection{Training details}
\label{sec:train}
We use the AdamW optimizer \citep{loshchilov2017decoupled} with weight decay $0.01$ and an initial learning rate $0.001$. We train the network with supervised training data for 200 epochs. The last 50 epochs use a learning rate of $0.0001$. We set the batch size to 256. The whole training procedure took around 70 hours to finish on 5 Tesla V100 GPU cards.


\end{document}